# Joint Information Preservation for Heterogeneous Domain Adaptation

Peng Xu, Zhaohong Deng, Kup-Sze Choi, Jun Wang, Shitong Wang
School of Digital Media, Jiangnan University, China
6171610015@stu.jiangnan.edu.cn

## Abstract

Domain adaptation aims to assist the modeling tasks of the target domain with knowledge of the source domain. The two domains often lie in different feature spaces due to diverse data collection methods, which leads to the more challenging task of *heterogeneous domain adaptation* (HDA). A core issue of HDA is how to preserve the information of the original data during adaptation. In this paper, we propose a joint information preservation method to deal with the problem. The method preserves the information of the original data from two aspects. On the one hand, although paired samples often exist between the two domains of the HDA, current algorithms do not utilize such information sufficiently. The proposed method preserves the paired information by maximizing the correlation of the paired samples in the shared subspace. On the other hand, the proposed method improves the strategy of preserving the structural information of the original data, where the local and global structural information are preserved simultaneously. Finally, the joint information preservation is integrated by distribution matching. Experimental results show the superiority of the proposed method over the state-of-the-art HDA algorithms.

## 1 Introduction

Domain Adaptation (DA) utilizes abundant labeled data from the source domain to assist the modeling tasks of a different but related target domain with only few (or no) labeled data [Csurka *et al.*, 2017; Weiss *et al.*, 2016]. According to the consistency of the feature spaces between the two domains, DA can be divided into homogeneous DA and *heterogeneous* DA (HDA). Based on the availability of labeled data in the target domain, DA can be further divided into unsupervised and semi-supervised DA [Day and Khoshgoftaar, 2017].

There are mainly three strategies in the existing DA algorithms, i.e. instance-based adaptation [Aljundi *et al.*, 2015], model-based adaptation [Long *et al.*, 2014] and feature-based adaptation [Sun *et al.*, 2016; Yan *et al.*, 2018]. Attributed to the flexibility of representation learning and classifier selection, feature-based adaptation methods have attracted extensive attention. Generally speaking, the basic idea of feature-based adaptation consists of two steps: one is to minimize the distance of the projected data between the two domains in the shared subspace; the other is to preserve the information of the original data during adaptation. Hence, the differences among the existing feature-based adaptation methods can be divided into three aspects, i.e., the way to construct the shared subspace, the way to measure the distance of the two domains and the way to preserve the information, which are detailed in the following.

For the first aspect, construction of the shared subspace can be achieved through symmetric feature transformation or asymmetric feature transformation. The data of the two domains are transformed into an intermediate shared subspace for the former [Zhong *et al.*, 2009] while the data are usually transformed from one domain to the other for the latter [Fernando *et al.*, 2014]. For the second aspect, measurement of the distance between the two domains can be performed using the *maximum mean discrepancy* (MMD) [Li *et al.*, 2018], Kullback-Leibler divergence [Zhuang *et al.*, 2015] or the Wasserstein distance [Shen *et al.*, 2018]. For the third aspect, the existing methods mainly makes various manifold assumptions of the original data for information preservation, such as principal component analysis [Long *et al.*, 2013], locality preserving projections [Wang and Mahadevan, 2011] and discriminative local alignment [Si *et al.*, 2010].

Although these existing methods have achieved good performance, they suffer from two shortcomings from the aspect of information preservation. First, in many scenes of HDA, paired samples often exist between the two domains. Taking the image-text HDA classification as an example. Despite of abundant paired samples on the Internet where the image and text on the same page are matched, the existing methods hardly utilize such paired information. Second, the structural information preservation methods adopted by the existing methods is so specific that either the local or the global manifold is adopted, thereby reducing the generalization performance.

To overcome the above challenges, a feature-based adaptation algorithm called *joint information preservation* (JIP) is proposed in the paper. It aims to tackle the tasks under the scene of semi-supervised HDA. Unlike the existing methods that only adopt simple strategies the for information preservation during the construction of the shared subspace, the proposed method incorporates the paired information and the structural information preservation into a unified HDA framework. Specifically, JIP constructs a shared subspace by symmetric transformation and the transformed matrices are optimized through three aspects: 1) the distribution distance of the two domains is minimized in the shared subspace for adaptation, 2) the paired information is preserved by maximizing the correlation of paired samples, and 3) the structural information is preserved with local and global manifold methods simultaneously.

The contributions of this paper are highlighted below:

- The paired information and more robust structural information are jointly preserved, which alleviates the problem of information loss during adaptation.

- A HDA framework is proposed by integrating joint information preservation and distribution matching. Optimization is then formulated as the problem of generalized eigenvalue decomposition.

- Superiority of the proposed method over state-of-the-art methods is verified with experiments of image classification, action recognition and multimedia retrieval.

## 2 Related Work

In this section, classical HDA algorithms are first reviewed, followed by researches related to the proposed method.

Among feature-based HDA algorithms, [Shi *et al.*, 2010] first proposes a classical HDA framework and its implementation, the Hemap, which adopts symmetric transformation to construct the shared subspace and the transformed matrices are optimized by minimizing the distance of the projected data. Algorithm DAMA proposed in [Wang and Mahadevan, 2011] introduces manifold alignment into HDA to preserve manifold within each domain and align the manifolds with labels between the domains. Besides, asymmetric transformation is introduced into HDA to yield the algorithm ARC-t [Kulis *et al.*, 2011]. In model-based HDA methods, MMDT is similar to ARC-t in that it adopts the asymmetric transformation and integrates the optimization of the large-margin model to obtain an adaptive SVM [Hoffman *et al.*, 2013]. As an extension of MMDT, the model-based method HFA conducts feature augmentation after feature transformation and increases the similarity of the samples within each domain by the augmented feature [Li *et al.*, 2014]. Unlike the previous algorithms that adapt the transformed feature, SHFR adapts the parameters of the trained classifiers [Zhou *et al.*, 2014]. Instance-based adaptation is another important strategy for HDA. CDLS proposed in [Tsai *et al.*, 2016] assigns each sample with a weigh during the optimization. The samples with nonzero weights are selected as the landmarks and the adaptation is based on these landmarks. [Li *et al.*, 2018] proposes the general HDA framework TIT to integrate distribution matching, manifold learning, sample weighting and feature selection. With the booming of neural networks, deep learning based HDA algorithms are also proposed. TNT proposed in [Chen *et al.*, 2016] designs a neural network based architecture for HDA. [Wang *et al.*, 2017] utilizes the autoencoder to develop a deep HDA algorithm for better information preservation. A comprehensive review of HDA can be found in [Csurka *et al.*, 2017; Day *et al.*, 2017].

There have been some researches considering paired samples for DA. They can be divided into two categories. The first category is multi-view DA that assumes that the data in both domains have multiple views [Hoffman *et al.*, 2012; Yang and Gao, 2013]. Multi-view DA aims to assist the learning tasks of the target domain by utilizing the labeled and multi-view samples in the source domain, which is different from the scene concerned in this paper. The second category considers that there is only single view in each domain and that paired samples exist across the two domains. The work of the paper falls into this category. [Yeh *et al.*, 2014] first leverages paired samples across domains for HDA. The *canonical correlation analysis* (CCA) is used to construct a correlated subspace for adaptation and CTSVM is proposed by integrating the classifier optimization. Similar to the basic idea of CTSVM, DCA proposed in [Yan *et al.*, 2017] adopts a different method alternating direction method of multipliers to optimize the problem. RSP-KCCA proposed in [Mehrkanoon and Suykens, 2018] kernelizes CCA and formulates the problem into a least square SVM. There is a common problem among CTSVM, DCA and RSP-KCCA, i.e., the construction of the shared subspace for adaptation totally depends on the paired samples. If only few paired samples are available, the adaptability of the shared subspace will be decreased severely.

## 3 Joint Information Preservation

In this section, the details of the propose method is presented under the settings of HDA.

### 3.1 Notations and Problem Formulation

In the scene of semi-supervised HDA, the feature spaces of the source domain and target domain are different. The task is to enhance the performance of the target domain by leveraging the data of the source domain.

Denote the data of the source domain $\mathbf{X}_S = \{\mathbf{x}_i^s\}_{i=1}^{n_s}$ with the corresponding labels $\mathbf{Y}_S = \{y_i^s\}_{i=1}^{n_s}$ and the data of the target domain $\mathbf{X}_T = \{\mathbf{x}_i^s\}_{i=1}^{n_t}$ with the corresponding pseudo labels $\hat{\mathbf{Y}}_T = \{\hat{y}_i^t\}_{i=1}^{n_t}$, where $\mathbf{X}_S \in R^{d_s \times n_s}$ and $\mathbf{X}_T \in R^{d_t \times n_t}$, $d_s$ and $d_t$ represent the feature dimensions of the two domains, $n_s$ and $n_t$ represent the number of samples of the two domains. Assume that the number of paired samples is $n_p$ and $n_p \leq \min\{n_s, n_t\}$, the paired samples can be represented as $\mathbf{X}_{SP} = \{\mathbf{x}_i^s\}_{i=1}^{n_p}$ and $\mathbf{X}_{TP} = \{\mathbf{x}_i^t\}_{i=1}^{n_p}$. Since the labels of the paired samples in the two domains are shared, the labels of the paired samples of two domains are represented as $\mathbf{Y}_{SP} = \{y_i^s\}_{i=1}^{n_p}$ and $\mathbf{Y}_{TP} = \{y_i^t\}_{i=1}^{n_p}$. Following the basic idea of feature-based adaptation, the objective function of the proposed method can be represented as follows,

$$\min_{\phi} \underbrace{D(\mathbf{X}_S, \mathbf{X}_T \mid \phi)}_{\text{distribution distance}} + \underbrace{L(\mathbf{X}_{SP}, \mathbf{X}_{TP} \mid \phi) + L(\mathbf{X}_S, \mathbf{X}_T \mid \phi)}_{\text{joint paired and structural information loss}}, \quad (1)$$

where $\phi$ denotes the feature transformation to construct the shared subspace. The first term of (1) is to minimize the distribution distance of the two domains in the shared subspace and the second term is to jointly preserve the paired information and the structural information.

## 3.2 Distribution Matching

In general, the basic step of feature-based HDA is to match the distributions of the two domains in the shared subspace. In the paper, the strategy of joint distribution adaptation (JDA) [Long *et al.*, 2013] is adopted. However, unlike the shared transformation matrix that is adopted in JDA, two transformation matrices $\mathbf{A}$ and $\mathbf{B}$ are used in proposed JIP to bridge the heterogeneous features spaces. Here, $\mathbf{A} \in R^{d_s \times m}$, $\mathbf{B} \in R^{d_t \times m}$ and $m$ is the dimension of the shared subspace. JDA jointly adapts both the marginal and the conditional distributions using MMD. The optimization of JDA in the scene of the HDA can be formalized as follows,

$$\min_{\mathbf{A},\mathbf{B}} \left\| \frac{1}{n_s} \sum_{i=1}^{n_s} \mathbf{A}^T \mathbf{x}_i^s - \frac{1}{n_t} \sum_{j=1}^{n_t} \mathbf{B}^T \mathbf{x}_j^t \right\|_F^2, \quad (2a)$$

$$\min_{\mathbf{A},\mathbf{B}} \sum_{c=1}^{C} \left\| \frac{1}{n_s^c} \sum_{y_i^s=c} \mathbf{A}^T \mathbf{x}_i^s - \frac{1}{n_t^c} \sum_{\hat{y}_j^t=c} \mathbf{B}^T \mathbf{x}_j^t \right\|, \quad (2b)$$

where (2a) represents the adaptation for the marginal distribution and (2b) represents the adaptation for the conditional distribution, $C$ is the number of classes, $n_s^c$ and $n_t^c$ represent the number of samples in class $c$ for the source and the target domains respectively. Similar to JDA, an iterative pseudo label refinement strategy is adopted, which will be detailed in Algorithm 1. With $\mathbf{W}^T = [\mathbf{A}^T, \mathbf{B}^T]$, the following objective can be obtained by unifying (2a) and (2b),

$$\min_{\mathbf{W}} \mathrm{Tr}(\mathbf{W}^T \mathbf{X}(\mathbf{M}_0 + \sum_{i=1}^{C} \mathbf{M}_c) \mathbf{X}^T \mathbf{W}), \quad (3)$$

where $\mathbf{X} = [\mathbf{X}_S, \mathbf{0}_{d_s \times n_t}; \mathbf{0}_{d_t \times n_s}, \mathbf{X}_T]$. Here, $\mathbf{M}_0$ and $\mathbf{M}_c$ are the MMD matrices with $c = 1, 2, ..., C$, that are used for marginal and conditional distributions matching. They can be computed in the same as those in JDA [Long *et al.*, 2013].

## 3.3 Paired Information Preservation

To preserve the paired information, we use CCA to maximize the correlation of the paired samples. CCA aims to find a pair of projected vectors $\mathbf{a} \in R^{d_s \times 1}$ and $\mathbf{b} \in R^{d_t \times 1}$ to maximize the correlation between the projected data $\mathbf{a}^T \mathbf{X}_{SP}$ in the source domain and the projected data $\mathbf{b}^T \mathbf{X}_{TP}$ in the target domain. The objective function of CCA is given by

$$\max_{\mathbf{a},\mathbf{b}} \frac{\mathbf{a}^T \mathbf{X}_{SP} \mathbf{H}_P \mathbf{X}_{TP}^T \mathbf{b}}{\sqrt{\mathbf{a}^T \mathbf{X}_{SP} \mathbf{H}_P \mathbf{X}_{SP}^T \mathbf{a}} \sqrt{\mathbf{b}^T \mathbf{X}_{TP} \mathbf{H}_P \mathbf{X}_{TP}^T \mathbf{b}}}, \quad (4)$$

where $\mathbf{H}_P$ denotes centering matrix which can simplify the calculation of covariance and variance in (4). Denote the identity matrix as $\mathbf{I}_P \in R^{n_p \times n_p}$ and column vector with all ones as $\mathbf{1}_P \in R^{n_p \times 1}$, then $\mathbf{H}_P = \mathbf{I}_P - (1/n_p)\mathbf{1}_P \mathbf{1}_P^T$.

By optimizing (4), only one pair of projected vectors $\mathbf{a}$ and $\mathbf{b}$ can be obtained and the projected subspace lies in one-dimensional space. To span the projected data in higher dimensional space, it is necessary to jointly optimize a group of correlation coefficients, where more than one pair of projected vectors $\mathbf{A} = [\mathbf{a}_1, \mathbf{a}_2, ..., \mathbf{a}_m]$ and $\mathbf{B} = [\mathbf{b}_1, \mathbf{b}_2, ..., \mathbf{b}_m]$ can be obtained. Then the optimization problem in (4) for a group of correlation coefficients can be formulated as follows,

$$\max_{\mathbf{A},\mathbf{B}} \mathrm{Tr}(\mathbf{A}^T \mathbf{X}_{SP} \mathbf{H}_{n_p} \mathbf{X}_{TP}^T \mathbf{B})$$
$$\text{s.t.} \quad \mathbf{A}^T \mathbf{X}_{SP} \mathbf{H}_{n_p} \mathbf{X}_{SP}^T \mathbf{A} = \mathbf{I}, \quad \mathbf{B}^T \mathbf{X}_{TP} \mathbf{H}_{n_p} \mathbf{X}_{TP}^T \mathbf{B} = \mathbf{I}. \quad (5)$$

Rescaling of the projected vectors will not affect the solution of (4), which is the reason of derived constrained optimization of (5). The most commonly used method to optimize (5) is Lagrange multipliers [Hardoon et al., 2014], and the projected matrices $\mathbf{A}$ and $\mathbf{B}$ can be obtained *sequentially*.

In the proposed method, we want to integrate the paired information preservation into the framework of distribution matching. Hence, the projected matrices need to be optimized *simultaneously* rather than sequentially. Given the paired samples in the two domains, $\mathbf{X}_{SP} \mathbf{H}_{n_p} \mathbf{X}_{SP}^T$ and $\mathbf{X}_{TP} \mathbf{H}_{n_p} \mathbf{X}_{TP}^T$ are both fixed. Therefore, the effect of equality constraints in (5) is to limit the value of the projected vectors so that their directions can be optimized. To simultaneously optimize the projected matrices, (5) is formulated as

$$\max_{\mathbf{A},\mathbf{B}} \mathrm{Tr}(\mathbf{A}^T \mathbf{X}_{SP} \mathbf{H}_{n_p} \mathbf{X}_{TP}^T \mathbf{B} + \mathbf{B}^T \mathbf{X}_{TP} \mathbf{H}_{n_p} \mathbf{X}_{SP}^T \mathbf{A}). \quad (6)$$

The constraint condition for $\mathbf{A}$ and $\mathbf{B}$ will be discussed in Section 3.5. Since $\mathbf{W}^T = [\mathbf{A}^T, \mathbf{B}^T]$, (6) can be expressed as

$$\max_{\mathbf{W}} \mathrm{Tr}(\mathbf{W}^T \mathbf{C} \mathbf{W}),$$
$$\mathbf{C} = [\mathbf{0}_{d_s \times d_t}, \mathbf{X}_{SP} \mathbf{H}_{n_p} \mathbf{X}_{TP}^T; \mathbf{X}_{TP} \mathbf{H}_{n_p} \mathbf{X}_{SP}^T, \mathbf{0}_{d_t \times d_s}], \quad (7)$$

where $\mathbf{C}$ is named as the correlation matrix and the paired information preservation is formulated with (7) ultimately.

## 3.4 Structural Information Preservation

To preserve the structural information of the original data more efficiently, the proposed JIP simultaneously adopts the local and global manifold methods. Discriminative manifold methods are adopted to utilize the labels of the data.

**Local Structure Preservation**

To preserve the local manifold structure of the data, the *locality preserving projections* (LPP) [He and Niyogi, 2003] is introduced. LPP is a type of linear approximation of Laplacian eigenmaps [Belkin and Niyogi, 2001], where the neighborhood structure of the original samples is still remained in the shared subspace. The objective function of LLP is

$$\min_{\mathbf{A},\mathbf{B}} \frac{1}{2} \sum_{i,j=1}^{n_s+n_t} \|\mathbf{z}_i - \mathbf{z}_j\|^2 \mathbf{W}_{ji}^L, \quad \mathbf{z}_l = [\mathbf{W}^T \mathbf{X}]_l, \quad (8)$$

where $l = 1, 2, ..., (n_s + n_t)$ is the index of the projected samples. $\mathbf{W}^L$ is the adjacency matrix, where $\mathbf{W}_{ij}^L$ is the distance measure between the samples $\mathbf{x}_i$ and $\mathbf{x}_j$. Define $\mathbf{D}$ as a diagonal matrix with $\mathbf{D}_{ii} = \sum_i \mathbf{W}_{ij}^L$ and $\mathbf{L}$ as the Laplacian matrix with $\mathbf{L} = \mathbf{D} - \mathbf{W}^L$. The objective function in the form of the trace of the matrix can be derived from (8) as follows.

**Algorithm 1** JIP
**Input**: Data of two domains; dimension of the shared subspace; regularization parameters; number of iterations.
**Output**: Labels of the unlabeled data in the target domain.
**JIP Procedure**:
1: Label the unlabeled data using pre-trained classifier with the labeled data in the target domain.
2: Calculate $\mathbf{M}_0$ and $\mathbf{C}$ in (3) and (7) respectively.
3: **For** $t \leftarrow 1,2,...,T$ **do**
4:   Update $\mathbf{M}_c$ in (3).
5:   Update $\mathbf{L}$ in (9).
5:   Update $\mathbf{S}_b$ and $\mathbf{S}_w$ in (10a) and (10b)
6:   Update $\mathbf{W}$ based on (15) using generalized eigenvalue decomposition.
7:   Calculate the data in the shared subspace.
8:   Train the classifier using the new data, update the pseudo labels for the data in the target domain $\hat{\mathbf{Y}}_T = \{\hat{y}_i^t\}_{i=1}^{n_t}$.
9: **end for**

$$\min_{\mathbf{W}} \text{tr}(\mathbf{W}^T \mathbf{X} \mathbf{L} \mathbf{X}^T \mathbf{W}) \quad (9)$$

The Laplacian matrix $\mathbf{L}$ can be calculated with a given $\mathbf{W}^L$ which is constructed by the distance between each pair of samples. There are many methods to measure the distance between samples, such as Euclidean distance, cosine similarity, local neighborhood relationship and label information. To leverage the label information, $\mathbf{W}^L$ is constructed using cosine distance in a discriminative manner [Li *et al.*, 2018] in this paper.

**Global Structure Preservation**
Besides the local structure, preservation of the global structural information is also important for unknown data structure. In this paper, linear discriminative analysis is adopted to preserve the global structural information, i.e., minimizing the within-class scatter and maximizing the between-class scatter. The objective function is given by

$$\max_{\mathbf{W}} \text{Tr}(\mathbf{W}^T \mathbf{S}_b \mathbf{W}), \quad \mathbf{S}_b = [\mathbf{S}_{sb}, \mathbf{0}_{d_s \times d_t}; \mathbf{0}_{d_t \times d_s}, \mathbf{S}_{tb}], \quad (10a)$$

$$\min_{\mathbf{W}} \text{Tr}(\mathbf{W}^T \mathbf{S}_w \mathbf{W}), \quad \mathbf{S}_w = [\mathbf{S}_{sw}, \mathbf{0}_{d_s \times n_t}; \mathbf{0}_{d_t \times n_s}, \mathbf{S}_{tw}], \quad (10b)$$

where $\mathbf{S}_b$ and $\mathbf{S}_w$ are the between-class scatter matrix and within-class scatter matrix respectively. $\mathbf{S}_{sb}$ and $\mathbf{S}_{sw}$ are the scatter matrices for the data in the source domain, $\mathbf{S}_{tb}$ and $\mathbf{S}_{tw}$ are the scatter matrices for the data in the target domain. They are calculated as follows,

$$\mathbf{S}_{sw} = \sum_{i=1}^{C} \mathbf{X}_S^i \mathbf{H}_S^i (\mathbf{X}_S^i)^T, \quad (11a)$$

$$\mathbf{S}_{sb} = \sum_{i=1}^{C} m_s^i (\boldsymbol{\mu}_s^i - \boldsymbol{\mu}_s)(\boldsymbol{\mu}_s^i - \boldsymbol{\mu}_s)^T, \quad (11b)$$

$$\mathbf{S}_{tw} = \sum_{i=1}^{C} \mathbf{X}_T^i \mathbf{H}_T^i (\mathbf{X}_T^i)^T, \quad (11c)$$

$$\mathbf{S}_{tb} = \sum_{i=1}^{C} m_t^i (\boldsymbol{\mu}_t^i - \boldsymbol{\mu}_t)(\boldsymbol{\mu}_t^i - \boldsymbol{\mu}_t)^T, \quad (11d)$$

where the subscripts $S$, $s$ and $T$, $t$ denote the data in the source domain and the target domain respectively. $\mathbf{X}_S^i$ and $\mathbf{X}_T^i$ are the data matrices of the $i$th class; $m_s^i$ and $m_t^i$ are the number of samples belonging to the $i$th class. $\mathbf{H}_S^i$ and $\mathbf{H}_T^i$ are the centering matrices for the samples belonging to the $i$th class. The calculation of $\mathbf{H}_S^i$ and $\mathbf{H}_T^i$ is similar that of $\mathbf{H}_P$ in (4), the only difference is that $n_p$ is replaced by $m_s^i$ or $m_t^i$. $\boldsymbol{\mu}_s^i$ and $\boldsymbol{\mu}_t^i$ are the mean of the samples belonging to the $i$th class. $\boldsymbol{\mu}_s$ and $\boldsymbol{\mu}_t$ represent the mean of all the samples in the source and target domains respectively.

### 3.5 Overall Objective Function and Optimization

Through the integration of the objective functions (3), (7), (9) and (10), the overall objective function can be obtained by introducing the regularization parameters $\alpha$, $\beta$ and $\lambda$ to balance among the preservation of local structure, paired information and global structure.

$$\min_{\phi} \underbrace{D(\mathbf{X}_S, \mathbf{X}_T \mid \phi)}_{\text{distribution distance}} + \underbrace{L(\mathbf{X}_{SP}, \mathbf{X}_{TP} \mid \phi) + L(\mathbf{X}_S, \mathbf{X}_T \mid \phi)}_{\text{joint paired and structural information loss}}$$

$$= \min_{\mathbf{W}} \text{Tr} \left( \frac{\mathbf{W}^T (\mathbf{X}(\mathbf{M}_0 + \sum_{i=1}^{C} \mathbf{M}_c + \alpha \mathbf{L}) \mathbf{X}^T + \lambda \mathbf{S}_w) \mathbf{W}}{\mathbf{W}^T (\beta \mathbf{C} + \lambda \mathbf{S}_b) \mathbf{W}} \right) \quad (12)$$

Note that the rescaling of $\mathbf{W}$ does not affect the solution of (12). The denominator of (12) is treated as the constraint condition such that the optimization has a unique solution. In the meanwhile, the remaining problem in (6) is tackled, where the temporal discarding constraint condition for the projected matrices in (5) is added. Hence, the problem is transformed into the following objective function,

$$\min_{\mathbf{W}} \text{Tr}(\mathbf{W}^T (\mathbf{X}(\mathbf{M}_0 + \sum_{i=1}^{C} \mathbf{M}_c + \alpha \mathbf{L}) \mathbf{X}^T + \lambda \mathbf{S}_w) \mathbf{W})$$
$$\text{s.t. } \text{Tr}(\mathbf{W}^T (\beta \mathbf{C} + \lambda \mathbf{S}_b) \mathbf{W}) = 1 \quad (13)$$

Using Lagrange function, (13) can be optimized to give (14) as follows,

$$L = \text{Tr}(\mathbf{W}^T (\mathbf{X}(\mathbf{M}_0 + \sum_{i=1}^{C} \mathbf{M}_c + \alpha \mathbf{L}) \mathbf{X}^T + \lambda \mathbf{S}_w) \mathbf{W}) + \text{Tr}((\mathbf{I} - \mathbf{W}^T (\beta \mathbf{C} + \lambda \mathbf{S}_b) \mathbf{W}) \boldsymbol{\Phi}) \quad (14)$$

where $\boldsymbol{\Phi} = \text{diag}(\varphi_1, \varphi_2, \cdots, \varphi_m)$ is the Lagrange multipliers, and $m$ is the dimension of the shared subspace. By setting $\partial L / \partial \mathbf{W} = 0$, the following equation is obtained.

$$(\mathbf{X}(\mathbf{M}_0 + \sum_{i=1}^{C} \mathbf{M}_c + \alpha \mathbf{L}) \mathbf{X}^T + \lambda \mathbf{S}_w) \mathbf{W} = (\beta \mathbf{C} + \lambda \mathbf{S}_b) \mathbf{W} \boldsymbol{\Phi} \quad (15)$$

Hence, the optimization problem in (12) is transformed into the problem of generalized eigenvalue decomposition in (15). Finding the optimal $\mathbf{W}$ is then reduced to solving (15) for the $m$ smallest eigenvalues and the corresponding eigenvectors constitute the projected matrix $\mathbf{W}$. The algorithm flowchart is illustrated in Algorithm 1.

## 4 Experiments

### 4.1 Datasets

The proposed method is evaluated on three datasets which are Caltech-Office, IXMAS and WIKI.

Caltech-Office is an image classification dataset composed of Caltech and Office. The Office dataset, containing 31 classes, comes from three different sources, i.e., AMAZON (A), Webcam (W) and DSLR (D). The Caltech (C) dataset contains 256 classes. In the experiments, the four different sources A, W, D and C are treated as four small datasets, and ten common classes of these four datasets are used. Two types of features are extracted for all the images, i.e., the SURF and DeCAF features, in a way similar to that in [Tsai et al., 2016]. These two types of features are regarded as two views. To construct HDA tasks for the experiments, the views are further regarded as the source domain and the target domain respectively. Eight tasks are thus constructed as shown in Table 1. Taking the A-D2S as an example, it represents the adaptation from the source domain DeCAF to the target domain SURF on dataset A.

IXMAS is an action recognition dataset containing eleven classes. There are 36 samples for each class of action. The actions are captured by five cameras and each camera is treated as a view. Similar to the processing in [Mehrkanoon and Suykens, 2018], the samples are transformed into 1000-dimension vectors. In the experiments, samples from any two cameras are used to construct HDA tasks. Since a camera can be treated as the source domain or the target domain, 20 tasks can be constructed with five cameras.

WIKI is an image-text dataset, where each sample contains an image and the corresponding text description. As the way in [Mehrkanoon and Suykens, 2018], the images are represented as 128-dimension vectors using the method Scale Invariant Feature Transforms; the texts are represented as 10-dimension vectors using the Latent Dirichlet Allocation. In the experiments, five classes are selected, each containing 100 samples. The image and text can both be treated as the source domain or the target domain. Therefore, two HDA tasks img2txt and txt2img are constructed, where the img2txt represents adaptation from image to text and vice versa.

### 4.2 Experimental Settings

In the experiments, seven algorithms are adopted to compare with the proposed JIP. The baseline method is SVM$t$ which trains SVM by using only the labeled data in the target domain. The other six methods are all state-of-the-art HDA methods, including MMDT [Hoffman et al., 2013], CTSVM [Yeh et al., 2014], semi-supervised HFA (SHFA) [Li et al., 2014], CDLS [Tsai et al., 2016], TNT [Chen et al., 2016], and TIT [Li et al., 2018]. For all the methods, the number of iterations is set to 5; the dimension of the shared subspace is set to 100; the optimal regularization parameters involved are searched from the set {0, 0.01, 0.1, 1, 10, 100}.

For the experiments on Caltech-Office and IXMAS datasets, 30% samples of each domain are selected to constitute paired samples. For the WIKI dataset, 10%, 20%, 30% and 40% paired samples are constructed for the experiments to demonstrate the performance of JIP with different proportions of paired samples. In the target domain, only paired samples are labeled and the rest are unlabeled.

### 4.3 Results Analysis

The results of the experiments on the Caltech-Office dataset are shown in Table 1. The proposed method achieves the best or competitive performance for most of the tasks on the Caltech-Office dataset. It even ranks first when considering the performance on the eight tasks on average.

The results on the IXMAS dataset are shown in Figure 1. It is obvious that the performance of the proposed method also exceeds that of all the other algorithms on the 20 tasks on average and achieves the highest accuracy 80.38%.

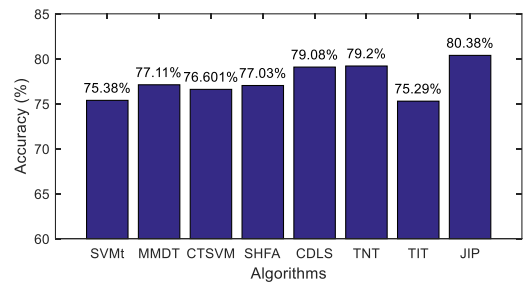

Fig. 1: Accuracy of algorithms on the IXMAS dataset

Table 1: Accuracy of algorithms on the Caltech-Office datasets (%)

| Tasks | SVM$t$ | MMDT | CTSVM | SHFA | CDLS | TNT | TIT | JIP |
|---|---|---|---|---|---|---|---|---|
| A-D2S | 66.77 | 66.77 | 69.00 | 70.64 | 67.96 | 71.41 | **72.88** | 71.54 |
| A-S2D | 93.29 | 92.70 | 93.44 | 93.44 | 93.59 | 94.13 | 93.59 | **95.23** |
| C-D2S | 55.27 | 54.51 | 52.10 | 55.02 | 53.37 | 56.17 | 57.30 | **58.07** |
| C-S2D | 87.47 | 86.66 | 86.66 | 89.58 | **90.47** | 88.75 | 89.32 | 90.22 |
| D-D2S | 47.27 | 62.73 | 57.27 | 62.73 | 63.64 | 64.73 | 66.36 | **67.27** |
| D-S2D | 80.00 | 95.45 | 88.18 | 89.09 | 94.55 | 94.31 | 93.64 | **98.18** |
| W-D2S | 69.08 | 74.88 | 76.33 | 74.88 | 75.85 | 79.46 | **80.19** | 79.23 |
| W-S2D | 94.69 | 95.17 | 92.75 | 94.20 | 95.65 | 95.78 | 96.62 | **96.62** |
| Average | 74.23 | 78.61 | 76.97 | 78.70 | 79.39 | 80.59 | 81.24 | **82.05** |

Table 2: Accuracy of algorithms on the WIKI datasets (%)

| | Tasks | SVM*t* | MMDT | CTSVM | SHFA | CDLS | TNT | TIT | JIP |
|---|---|---|---|---|---|---|---|---|---|
| 10% | img2txt | 92.67 | 94.89 | 88.89 | 91.56 | 94.67 | 93.91 | 92.67 | **96.00** |
| | txt2img | 41.11 | 35.56 | 44.22 | 44.22 | 40.22 | 46.02 | 44.44 | **47.56** |
| 20% | img2txt | 94.25 | 95.75 | 87.75 | 91.00 | 95.25 | 92.63 | 94.25 | **96.00** |
| | txt2img | 43.25 | 37.00 | 43.50 | 47.25 | 40.25 | 48.08 | 47.25 | **50.25** |
| 30% | img2txt | 93.57 | 96.29 | 91.14 | 94.57 | 94.57 | 94.34 | 96.00 | **96.57** |
| | txt2img | 49.14 | 37.14 | 46.29 | 52.86 | 48.29 | 51.17 | 50.57 | **54.57** |
| 40% | img2txt | 96.33 | 96.33 | 91.67 | 96.00 | 95.00 | 96.07 | **97.33** | 96.67 |
| | txt2img | 49.33 | 35.00 | 48.00 | 53.33 | 52.33 | 50.97 | 51.33 | **54.33** |
| Average | | 69.96 | 66.00 | 67.68 | 71.35 | 70.07 | 71.65 | 71.73 | **73.99** |

The results on the WIKI dataset are shown in Table 2, where different proportions of paired samples are selected for the experiments. It can be seen that the performance of the algorithm becomes better with the increasing of proportion of paired samples. JIP achieves the best result when considering the performance of the algorithms on all the tasks on average.

### 4.4 Model Analysis

The proposed method is further studied by analyzing the convergence and dimensionality, and also the effectiveness of the paired and the structural information preservation.

**Analysis of Convergence and Dimensionality**

Two important parameters of the proposed JIP is the number of iterations and the number of dimensions. Fig 2(a) and Fig 2(b) show the variation in accuracy of JIP on the Caltech-Office dataset with number of iterations and dimensions respectively. For the sake of clarity, the accuracy for each task is moved up or down on the whole, which dose not affect the trend analysis. As shown in Fig 2(a), the proposed method demonstrates good convergence. It can be seen from Fig 2(b) that the trend under different number of dimensions is different for different tasks, and highest accuracy does not necessarily occur at the highest dimensionality, i.e., 100 in the experiments. If all the other parameters are fixed, the number of dimensions varies from 10 to 100 with the step of 10, the average accuracy of the proposed method on the eight tasks is 82.30%.

**Effectiveness of Information Preservation**

The effectiveness of information preservation of the proposed method is analyzed from four aspects. In Fig 3(a)-(c), the regularization parameters $\alpha$, $\beta$ and $\lambda$ corresponding to the respective terms in (12) are set to 0 or optimized value with the other parameters fixed. It can be seen that the accuracy increases for almost all the tasks when the parameters are non-zero, which demonstrates the effectiveness of the information preservation. In Fig 3(d), regularization parameters for the local and global structure preservation terms are both set to 0. The accuracy improvement in Fig 3(d) is significantly higher than that in Fig 3(b) and (c), which verifies the effectiveness of hierarchical structure preservation when compared to individual local or global structure preservation.

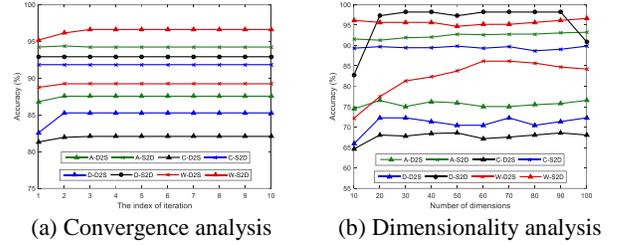

(a) Convergence analysis    (b) Dimensionality analysis

Fig. 2: Parameter analysis

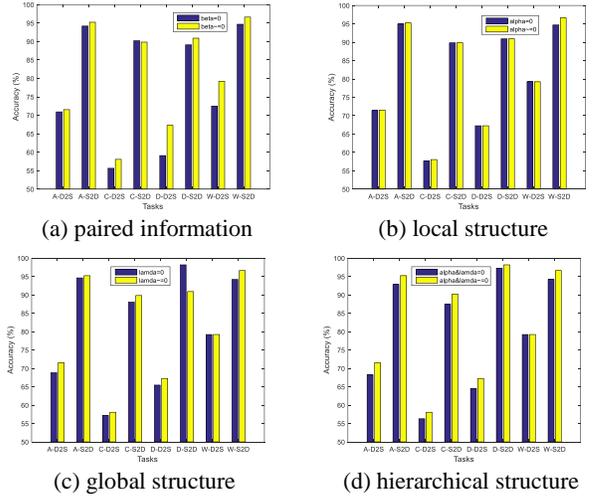

(a) paired information    (b) local structure

(c) global structure    (d) hierarchical structure

Fig. 3: Effectiveness analysis of information preservation

## 5 Conclusions

This paper purposes a new heterogeneous domain adaption method by preserving jointly the paired information and the structural information. Further, a HDA framework is proposed to integrate the joint information preservation with distribution matching that can alleviate the problem of information loss during adaptation effectively. A disadvantage is that the settings of the regularization parameters depend on grid search, which is time consuming. More adaptive strategy for setting the parameters will be explored in our future work.